\documentclass[10pt,twocolumn,letterpaper]{article}

\usepackage{iccv}
\usepackage{times}
\usepackage{epsfig}
\usepackage{graphicx}
\usepackage{amsmath}
\usepackage{amssymb}

\usepackage[accsupp]{axessibility}

\usepackage{booktabs}  
\usepackage{bm}        
\usepackage{enumitem}  

\usepackage[pagebackref=true,breaklinks=true,letterpaper=true,colorlinks,bookmarks=false]{hyperref}

\iccvfinalcopy

\newcommand{\paragraphskip}{\medskip}
\newcommand{\paragraphtitle}[1]{\noindent \textbf{#1}}

\newcommand{\tableskip}{\smallskip}

\setlist{parsep=0pt,itemsep=3pt,topsep=4pt,leftmargin=15pt}

\newcommand{\bb}{\bm{b}}
\newcommand{\bc}{\bm{c}}
\newcommand{\be}{\bm{e}}
\newcommand{\bk}{\bm{k}}

\newcommand{\bq}{\bm{q}}
\newcommand{\bu}{\bm{u}}
\newcommand{\bv}{\bm{v}}
\newcommand{\bx}{\bm{x}}
\newcommand{\by}{\bm{y}}
\newcommand{\bz}{\bm{z}}
\newcommand{\R}{\mathbb{R}}
\newcommand{\tbc}{\tilde{\bc}}
\newcommand{\tbe}{\tilde{\be}}
\newcommand{\tbk}{\tilde{\bk}}
\newcommand{\tbq}{\tilde{\bq}}
\newcommand{\tbv}{\tilde{\bv}}
\newcommand{\tbx}{\tilde{\bx}}
\newcommand{\tA}{\tilde{A}}

\def\iccvPaperID{5394}

\ificcvfinal\fi

\author{
    Matthew Dutson, Yin Li, and Mohit Gupta\\
    University of Wisconsin--Madison\\
    {\tt\small \{dutson,yin.li,mgupta37\}@wisc.edu}
}

\title{
    Eventful Transformers:\\
    Leveraging Temporal Redundancy in Vision Transformers
}

\begin{document}

\maketitle

\ificcvfinal\thispagestyle{empty}\fi

\begin{abstract}
Vision Transformers achieve impressive accuracy across a range of visual recognition tasks. Unfortunately, their accuracy frequently comes with high computational costs. This is a particular issue in video recognition, where models are often applied repeatedly across frames or temporal chunks. In this work, we exploit temporal redundancy between subsequent inputs to reduce the cost of Transformers for video processing. We describe a method for identifying and re-processing only those tokens that have changed significantly over time. Our proposed family of models, Eventful Transformers, can be converted from existing Transformers (often without any re-training) and give adaptive control over the compute cost at runtime. We evaluate our method on large-scale datasets for video object detection (ImageNet VID) and action recognition (EPIC-Kitchens 100). Our approach leads to significant computational savings (on the order of 2-4x) with only minor reductions in accuracy.
\end{abstract}

\section{Introduction}
\label{sec:introduction}

\begin{figure}
    \centering
    \includegraphics{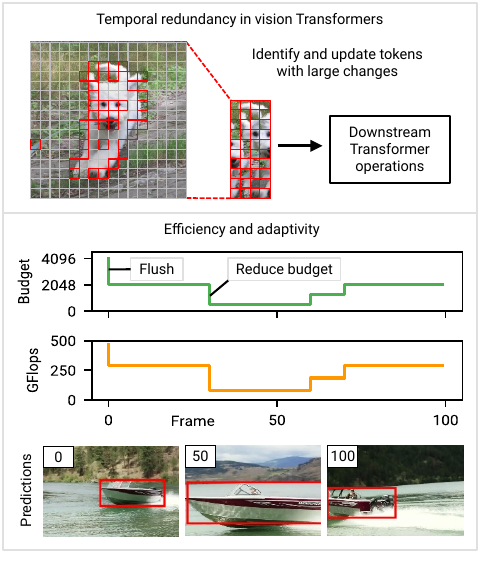}
    \caption{\textbf{Eventful Transformers.} Our method exploits temporal redundancy between subsequent model inputs. \textbf{(Top)} Within each Transformer block, we identify and update only those tokens with significant changes over time. Image: \cite{buzzfarmers2011}. \textbf{(Bottom)} In addition to improving efficiency, our method gives fine-grained control over the compute cost at runtime. ``Budget'' refers to parameter $r$ as described in Section~\ref{sec:policies}. ``Flush'' refers to the initialization of all tokens on the first time step. This example shows the ViTDet~\cite{liExploringPlain2022} object detection model on a video from the VID~\cite{russakovsky2015ImageNet} dataset.}
    \label{fig:teaser}
\end{figure}

Transformers, initially designed for language modeling~\cite{vaswaniAttentionAll2017}, have been recently explored as an architecture for vision tasks. Vision Transformers~\cite{dosovitskiyImageWorth2020} have achieved impressive accuracy across a range of visual recognition problems, attaining state-of-the-art performance in tasks including image classification~\cite{dosovitskiyImageWorth2020}, video classification~\cite{arnabViViTVideo2021,bertasiusSpacetimeAttention2021,fanMultiscaleVision2021}, and object detection~\cite{carionEndtoendObject2020,liExploringPlain2022,liuSwinTransformer2021,wangPyramidVision2021}.

One of the primary drawbacks of vision Transformers is their high computational cost. Whereas typical convolutional networks (CNNs) consume tens of GFlops per image~\cite{canzianiAnalysisDeep2017}, vision Transformers often require an order of magnitude more computation, up to hundreds of GFlops per image. In video processing, the large volume of data further amplifies these costs. High compute costs preclude vision Transformers from deployment on resource-constrained or latency-critical devices, limiting the scope of this otherwise exciting technology. In this paper, we present one of the first methods to use \emph{temporal redundancy between subsequent inputs} to reduce the cost of vision Transformers when applied to video data.

\paragraphskip
\paragraphtitle{Temporal redundancy.} Consider a vision Transformer that is applied frame-by-frame or clip-by-clip to a video sequence. This Transformer might be a simple frame-wise model (\eg, an object detector) or an intermediate step in some spatiotemporal model (\eg, the first stage of the factorized model from \cite{arnabViViTVideo2021}). Unlike in language processing, where one Transformer input represents a complete sequence, we consider Transformers applied to several distinct inputs (frames or clips) over time.

Natural videos contain significant temporal redundancy, with only slight differences between subsequent frames. Despite this fact, deep networks (including Transformers) are commonly computed ``from scratch'' on each frame. This approach is wasteful, discarding all potentially relevant information from previous inferences. Our key intuition is that we can reuse intermediate computations from earlier time steps to improve efficiency on redundant sequences.

\paragraphskip
\paragraphtitle{Adaptive inference.} For vision Transformers (and deep networks in general), the inference cost is typically fixed by the architecture. However, in real-world applications, the available resources may vary over time (\eg, due to competing processes or variations in power supply). As such, there is a need for models whose computational cost can be modified at runtime~\cite{mittalSurveyTechniques2016}. In this work, adaptivity is one of our primary design objectives; we design our method to allow real-time control over the compute cost. See Figure~\ref{fig:teaser} (bottom portion) for an example where we vary the compute budget throughout a video.

\paragraphskip
\paragraphtitle{Challenges and opportunities.} There are past works exploring temporal redundancy~\cite{dutsonEventNeural2022,habibianSkipconvolutionsEfficient2021,pargerDeltaCNNEndtoend2022} and adaptivity~\cite{mengARNetAdaptive2020,veitConvolutionalNetworks2018,xuSmartAdaptMultiBranch2022} for CNNs. However, these methods are generally incompatible with vision Transformers, owing to substantial architectural differences between Transformers and CNNs. Specifically, Transformers introduce a new primitive, self-attention, that does not conform to the assumptions of many CNN-based methods.

Despite this challenge, vision Transformers also represent a unique opportunity. In CNNs, it is difficult to translate sparsity improvements (\ie, the sparsity gained by considering temporal redundancy) into concrete speedups. Doing so requires imposing significant restrictions on the sparsity structure~\cite{habibianSkipconvolutionsEfficient2021} or using custom compute kernels~\cite{pargerDeltaCNNEndtoend2022}. In contrast, the structure of Transformer operations (centered on manipulating token vectors) makes it easier to translate sparsity into reduced runtime using standard operators.

\paragraphskip
\paragraphtitle{Eventful Transformers.} We propose Eventful Transformers, a new class of Transformer that leverages temporal redundancy between inputs to enable efficient, adaptive inference. The term ``Eventful'' is inspired by event cameras~\cite{brandli240x1802014,lichtsteiner128x1282008}, sensors that produce sparse outputs based on scene changes. Eventful Transformers track token-level changes over time, selectively updating the token representations and self-attention maps on each time step. Blocks in an Eventful Transformer include gating modules that allow controlling the number of updated tokens at runtime.

Our method can be applied to off-the-shelf models (generally without re-training) and is compatible with a wide range of video processing tasks. Our experiments demonstrate that Eventful Transformers, converted from existing state-of-the-art models, significantly reduce computational costs while largely preserving the original model's accuracy. We publicly release our code, which includes PyTorch modules for building Eventful Transformers. See our project page: \href{https://wisionlab.com/project/eventful-transformers/}{wisionlab.com/project/eventful-transformers}.

\paragraphskip
\paragraphtitle{Limitations.} We demonstrate wall-time speedups on both the CPU and GPU. However, our implementation (based on vanilla PyTorch operators) is likely sub-optimal from an engineering standpoint. With additional effort to reduce overhead (\eg, implementing a fused CUDA kernel for our gating logic), we are confident that the speedup ratios could be further improved. Our method also involves some unavoidable memory overheads. Perhaps unsurprisingly, reusing computation from previous time steps requires maintaining some tensors in memory. These memory overheads are relatively modest; see Section~\ref{sec:discussion} for further discussion.

\section{Related Work}
\label{sec:related_work}

\paragraphtitle{Efficient Transformers.} Several past works improve the efficiency of Transformers. Many of these methods focus on reducing the quadratic complexity of self-attention, often using low-rank or sparse approximations~\cite{childGeneratingLong2019,choromanskiRethinkingAttention2023,guoStarTransformer2022,jaszczurSparseEnough2021,katharopoulosTransformersAre2020,kitaevReformerEfficient2020,luSOFTSoftmaxfree2021,renCombinerFull2021,royEfficientContentbased2021,wangLinformerSelfattention2020}. In this work, we consider standard self-attention (with windowing in some cases). Our approach is orthogonal to the above methods.

\paragraphskip
\paragraphtitle{Selecting and summarizing tokens.} Some recent works improve the efficiency of vision Transformers by exploiting spatial redundancy within each input. Many of these methods prune or fuse tokens based on a salience measure~\cite{fayyazAdaptiveToken2022,goyalPoWERBERTAccelerating2020,liangEViTExpediting2022,panIAREDInterpretabilityaware2021,raoDynamicViTEfficient2021}. A notable example is the Adaptive Token Sampling (ATS) algorithm~\cite{fayyazAdaptiveToken2022}, which has an adaptive computation cost and does not require re-training. Other spatial redundancy methods include adaptive token pooling~\cite{bolyaTokenMerging2022,marinTokenPooling2021}, hierarchical pooling~\cite{panScalableVision2021}, learned tokenization~\cite{ryooTokenLearnerWhat2022}, and progressive token sampling~\cite{yueVisionTransformer2021}.

Unlike these works, which consider \emph{spatial} redundancy, our method targets \emph{temporal} redundancy. This makes our work complementary to these approaches. A single model can leverage both spatial and temporal redundancy by only updating tokens that are both salient \emph{and} not temporally repetitive. We illustrate this compatibility in our experiments by building a simple proof-of-concept model that combines temporal and spatial redundancy.

Another related work is Spatiotemporal Token Selection (STTS)~\cite{wangEfficientVideo2022}, which exploits spatiotemporal redundancy for video inference. STTS is intended for models with explicit temporal reasoning that take an entire video as input. In contrast, our method is designed for models that are repetitively applied to frames or clips. Compared to STTS, our method covers a wider range of architectures and tasks.

\paragraphskip
\paragraphtitle{Temporal redundancy between inputs.} There has been recent work on exploiting inter-frame temporal redundancy in CNNs~\cite{cavigelliCBinferChangebased2017,dutsonEventNeural2022,habibianSkipconvolutionsEfficient2021,pargerDeltaCNNEndtoend2022}. While we draw some inspiration from these methods, directly applying them to vision Transformers is not feasible due to significant architectural differences between CNNs and Transformers.

There is limited existing research on exploiting temporal redundancy between subsequent vision Transformers inputs. To our knowledge, the only past work in this area is the Spatiotemporal Gated Transformers (STGT) method~\cite{liSpatiotemporalGated2021}. There are two noteworthy differences between STGT and our work. Most notably, STGT only considers temporal redundancy within token-level operations (\eg, token-wise linear transforms), and not within the self-attention operator. Our method accelerates all major Transformer components, including self-attention. Further, STGT uses lossy gating logic that leads to accuracy degradation on long sequences with gradual changes. Our method avoids this issue by employing an improved, reference-based gating mechanism.

\paragraphskip
\paragraphtitle{Adaptive neural networks.} Many existing methods add adaptivity to deep CNNs~\cite{chinAdaScaleTowards2019,figurnovSpatiallyAdaptive2017,huangMultiScaleDense2018,mengARNetAdaptive2020,teerapittayanonBranchyNetFast2016,veitConvolutionalNetworks2018,wangSkipNetLearning2018,wuLiteEvalCoarsetoFine2019,xuSmartAdaptMultiBranch2022,yangResolutionAdaptive2020}. However, due to architectural differences (\eg, the use of relative position embeddings in Transformers), these methods (\eg, those based on input resizing) often do not translate to vision Transformers.

There has been some recent work on adaptive vision Transformers \cite{mengAdaViTAdaptive2022,wangNotAll2021,yinAViTAdaptive2022}. These works leverage redundancy within a single input, whereas we consider redundancy between inputs. Unlike our method, these approaches generally require re-training or fine-tuning the model.

\paragraphskip
\paragraphtitle{Efficient neural networks.} There is a substantial body of work on improving the efficiency of deep networks. Some works propose efficient CNN architectures \cite{howardMobileNetsEfficient2017,iandolaSqueezeNetAlexNetlevel2016,zhangShuffleNetExtremely2018}. Others use reduced-precision arithmetic \cite{courbariauxBinaryConnectTraining2015,hwangFixedpointFeedforward2014,rastegariXNORNetImageNet2016} or pruning \cite{hanLearningBoth2015,hassibiSecondOrder1992,lecunOptimalBrain1989,liPruningFilters2017}. Our method is loosely connected to pruning; it can be viewed as adaptively pruning redundant tokens on each time step.

\section{Background: Vision Transformers}
\label{sec:background}

In this section, we describe the basic elements of a vision Transformer (see \cite{dosovitskiyImageWorth2020} for more details) and define the notation we use throughout the rest of the paper.

A vision Transformer consists of a sequence of Transformer blocks. The input to each block is a list of $N$, $D$-dimensional token vectors; we denote this as $\bx \in \R^{N \times D}$. Before the first Transformer block, a vision Transformer maps each image patch to a token vector using a linear transform. Positional embedding~\cite{vaswaniAttentionAll2017} can be injected before the first block~\cite{dosovitskiyImageWorth2020} or at every block~\cite{liExploringPlain2022}.

\paragraphskip
\paragraphtitle{A Transformer block.} A Transformer block maps input $\bx \in \R^{N \times D}$ to output $\bz \in \R^{N \times D}$, according to
\begin{align}
    \by &= \text{MSA}(\text{LN}(\bx)) + \bx, \label{eq:msa} \\
    \bz &= \text{MLP}(\text{LN}(\by)) + \by, \label{eq:mlp}
\end{align}
where ``MSA'' denotes multi-headed self-attention. ``MLP'' is a token-wise multilayer perceptron with two layers and one GELU nonlinearity. ``LN'' denotes layer normalization.

\paragraphskip
\paragraphtitle{Multi-headed self-attention (MSA).} The self-attention operator first applies three linear transforms $W_q, W_k, W_v \in \R^{D \times D}$ to its input $\bx' = \text{LN}(\bx)$.
\begin{equation}
    \bq = \bx' W_q \qquad \bk = \bx' W_k \qquad \bv = \bx' W_v. \label{eq:qkv}
\end{equation}
$\bq$, $\bk$, and $\bv$ are the ``query,'' ``key,'' and ``value'' tensors, respectively. In practice, $W_q, W_k, W_v$ are often fused into a single transform $W_{qkv}=[W_q, W_k, W_v]$. These transforms may include a bias; we omit the bias here for brevity.

The self-attention operator then computes a normalized similarity matrix (attention matrix) $A \in \R^{N \times N}$ between the tokens of $\bq$ and $\bk$.
\begin{equation}
    A = \text{Softmax}\left(\bq \bk^T / \sqrt{D}\right). \label{eq:qk}
\end{equation}
Softmax normalization is applied along rows of the matrix.

The MSA output $\by'$ is an attention-weighted sum of the value tokens $\bv$, followed by a linear projection $W_p$.
\begin{equation}
    \by' = (A \bv) \, W_p. \label{eq:av_proj}
\end{equation}

\emph{Multi}-headed self-attention (as opposed to single-headed self-attention) splits $\bq$, $\bk$, and $\bv$ into $H$ tensors of shape $\R^{N \times (D / H)}$ and applies self-attention in parallel across these $H$ heads. Before applying $W_p$, the results of all heads are concatenated into a tensor with shape $\R^{N \times D}$.

\paragraphskip
\paragraphtitle{Windowed attention.}  Standard MSA has a complexity of $\mathcal{O}(N^2)$ (quadratic in the number of tokens). To reduce this cost, many vision Transformers adopt \emph{windowed} attention. Windowed attention constrains the attention computation to local windows. Information can be exchanged between windows by shifting the windows between blocks~\cite{liuSwinTransformer2021} or by interleaving global attention~\cite{liExploringPlain2022}.

\section{Eventful Transformers}
\label{sec:methods}

Our goal is to accelerate vision Transformers for video recognition, in the situation where a Transformer is applied repetitively across frames or chunks of frames (\eg, for video object detection or video action recognition, respectively). Our key idea is to exploit temporal redundancy by re-using computation from previous time steps. In this section, we describe how to modify Transformer blocks to add temporal redundancy awareness. 

In Section~\ref{sec:gating}, we present a token-gating module that monitors temporal changes and determines which tokens to update. In Section~\ref{sec:transformer}, we integrate our token gating logic into a Transformer block, creating a redundancy-aware Eventful Transformer block. In Section~\ref{sec:policies}, we explore policies for selecting which tokens to update.

\begin{figure}
    \centering
    \includegraphics{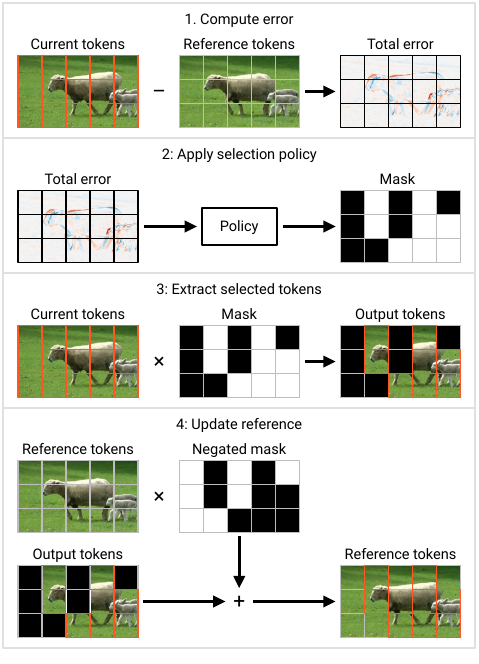}
    \caption{\textbf{Token Gating.} A gating module compares incoming tokens against a stored reference. If the difference between a token and its reference is large, then the token is selected to be updated. See Section~\ref{sec:policies} for details on selection policies. Images are from the VID~\cite{russakovsky2015ImageNet} dataset.}
    \label{fig:gate}
\end{figure}

\subsection{Token Gating: Detecting Redundancy}
\label{sec:gating}

In this subsection, we propose two modules: token gates and token buffers. These modules allow us to identify and update only those tokens that have changed significantly since their last update.

\paragraphskip
\paragraphtitle{Gate module.} A gate selects $M \leq N$ of its input tokens to send to downstream layers for re-computation. The gate maintains a set of \emph{reference tokens} in memory, which we denote as $\bu \in \R^{N \times D}$. The reference tensor contains the value of each token on the time step it was most recently updated. On each time step, tokens are compared against their references; those that deviate significantly from their reference are selected for an update.

Let $\bc \in \R^{N \times D}$ denote the current input to the gate. On each time step, we update the gate's state and determine its output according to the following procedure (see Figure~\ref{fig:gate}):
\begin{enumerate}
    \item Compute the total error $\be = \bu - \bc$.
    \item Apply a \emph{selection policy} to the error $\be$. A selection policy returns a binary mask $\bm{m}$ (equivalently, a list of token indices) indicating which $M$ tokens should be updated.
    \item Extract the tokens selected by the policy. In Figure~\ref{fig:gate}, we depict this as the product $\bc\times \bm{m}$; in practice, we achieve this with a ``gather'' operation along the first axis of $\bc$. We denote the gathered tokens as $\tbc \in \R^{M \times D}$. The gate returns $\tbc$ as its output.
    \item Update the references for selected tokens. In Figure~\ref{fig:gate}, we depict this as $\bu \leftarrow \be \times (\sim \bm{m}) + \bc \times \bm{m}$; in practice, we apply a ``scatter'' operation from $\tbc$ into $\bu$.
\end{enumerate}
On the first time step, the gate updates all tokens (initializing $\bu \leftarrow \bc$ and returning $\tbc = \bc$).

\begin{figure}
    \centering
    \includegraphics{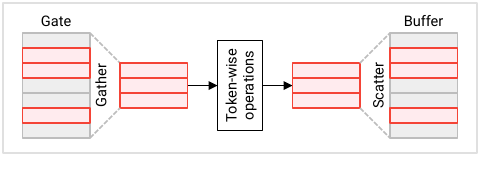}
    \caption{\textbf{Accelerating token-wise operations.} The gate reduces the number of active tokens from $N$ to $M$. Subsequent token-wise operations operate on a smaller tensor and therefore have a lower computational cost (proportional to $M$).}
    \label{fig:gathering}
\end{figure}

\paragraphskip
\paragraphtitle{Buffer module.} A buffer module maintains a state tensor $\bb \in \R^{N \times D}$ that tracks the most recent known value for each of its input tokens. When receiving a new input $f(\tbc) \in \R^{M \times D}$, the buffer scatters the tokens from $f(\tbc)$ into their corresponding locations in $\bb$. It then returns the updated $\bb$ as its output. See Figure~\ref{fig:gathering}.

We pair each gate with a subsequent buffer. One simple usage pattern is as follows. The gate output $\tbc \in \R^{M \times D}$ is passed to a series of token-wise operations $f(\tbc)$. The resulting tensor $f(\tbc) \in \R^{M \times D}$ is then passed to a buffer, which restores the full shape $\R^{N \times D}$.

\subsection{Building Redundancy-Aware Transformers}
\label{sec:transformer}

\begin{figure*}
    \centering
    \includegraphics{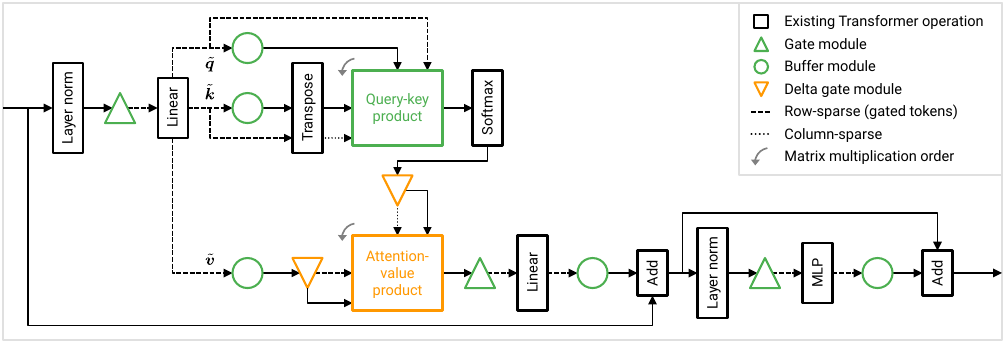}
    \caption{\textbf{An Eventful Transformer block.} To exploit temporal redundancy, we strategically apply token gating throughout the block and compute a modified, sparse self-attention update. Rectangles are standard Transformer components (see Section~\ref{sec:background}). For clarity, we have omitted some minor operations (\eg, scaling after the first matrix multiplication) from this figure.}
    \label{fig:overview}
\end{figure*}

In this subsection, we propose a modified Transformer block that exploits temporal redundancy. Figure~\ref{fig:overview} shows our design for an Eventful Transformer block. Our method accelerates token-wise operations (\eg, the MLP), as well as the query-key and attention-value multiplications (Equations~\ref{eq:qk} and \ref{eq:av_proj}, respectively).

\paragraphskip
\paragraphtitle{Token-wise operations.} Many of the operations in a Transformer block are token-wise, meaning they do not involve information exchange between tokens. These include the MLP and the linear transforms in the MSA. We can save computation in token-wise operations by skipping those tokens not selected by the gate. Due to token-wise independence, this does not change the result of the operation for the selected tokens. See Figure~\ref{fig:gathering}.

Specifically, we place a gate-buffer pair around each contiguous sequence of token-wise operations, including the $W_{qkv}$ transform (Equation~\ref{eq:qkv}), the $W_p$ transform (Equation~\ref{eq:av_proj}), and the MLP. Note that we add buffers before the skip connections (Equations~\ref{eq:msa} and \ref{eq:mlp}) to ensure that the tokens of the two addition operands are correctly aligned.

The cost of a token-wise operation is proportional to the number of tokens. A gate reduces the number of tokens from $N$ to $M$. This, in turn, reduces the computational cost of downstream token-wise operations by a factor of $N / M$.

\paragraphskip
\paragraphtitle{The query-key product.} We now consider the query-key product $B = \bq \bk^T$ (part of Equation~\ref{eq:qk}). Writing this matrix multiplication explicitly, we have
\begin{equation}
    B_{ij} = \sum_p \bq_{ip} \left( \bk^T \right)_{pj}.
\end{equation}
Element $B_{ij}$ needs to be updated if either (a) there is a change in the $i^\text{th}$ row of $\bq$, or (b) there is a change in the $j^\text{th}$ column of $\bk^T$. Due to the gate that we inserted before the $W_{qkv}$ transform (shown in Figure~\ref{fig:overview}), only some rows of $\bq$ and some columns of $\bk^T$ have changed. Therefore, we only need to recompute a subset of the elements of $B$.

Let $\tbx' \in \R^{M \times D}$ denote the output of the gate before the $W_{qkv}$ transform. We define $\tbq = \tbx' W_q$ and $\tbk = \tbx' W_k$ (following Equation~\ref{eq:qkv}). Let $\bq$ and $\bk$ denote the outputs of the $\tbq$ and $\tbk$ buffers (shown in Figure~\ref{fig:overview}). $\tbq$ contain $\tbk$ the subset of tokens from $\bq$ and $\bk$ that are being updated.

\begin{figure}
    \centering
    \includegraphics{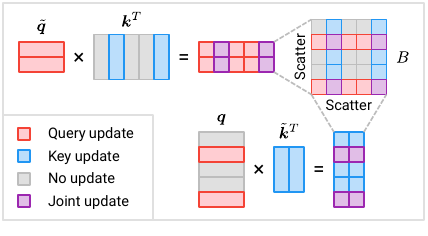}
    \caption{\textbf{The query-key product.} We reduce the cost of computing $B = \bq \bk^T$ by only updating a subset of its elements. We first compute changes induced by updated rows in $\bq$ (top-left), then compute changes induced by updated columns in $\bk^T$ (bottom).}
    \label{fig:matmul1}
\end{figure}

Figure~\ref{fig:matmul1} depicts our method for sparsely updating $B$. The product $\tbq \bk^T$ contains the elements of $B$ that need to be updated due to a change in $\tbq$. We compute $\tbq \bk^T$, then scatter the result row-wise into the old $B$ (the value of $B$ from the last time step). We use an analogous approach for the $\tbk$-induced updates; we compute $\bq \tbk^T$ and scatter the result column-wise into $B$.

The overall cost of these updates is $2 N M D$, compared to a cost of $N^2 D$ to compute $B$ from scratch. Note that the cost of our method is proportional to $M$, the number of tokens selected by the gate. We save computation when $M < N / 2$ (when we update fewer than half of the tokens).

The above method for updating $B$ involves some redundant computation. Some elements of the first scattered matrix $\tbq \bk^T$ are also present in the second matrix $\bq \tbk^T$. These overlapping elements are computed twice. Eliminating this redundancy would reduce the cost of our method to $N M D \leq N^2 D$. This could be achieved by removing the tokens in $\tbq$ from $\bq$ before computing $\bq \tbk^T$. We would then scatter the result by indexing along both axes of $B$. We leave this as an optimization for future implementations.

\paragraphskip
\paragraphtitle{The attention-value product.} We now describe a method for updating the attention-value product $A \bv$ (part of Equation~\ref{eq:av_proj}). Writing this multiplication explicitly, we have
\begin{equation}
    (A \bv)_{ij} = \sum_p A_{ip} \bv_{pj}.
\end{equation}
Because of the gate before the $W_{qkv}$ transform, only some rows (tokens) of $\bv$ change on each time step. However, there are some updated values in every \emph{column} of $\bv$. Therefore, every element of $A \bv$ will change on each time step. This means we cannot use the same strategy that we used for $B$, where we only updated some of the output elements.

Instead, we propose a delta-based update strategy. Let $A_o$ and $\bv_o$ denote the last known values for $A$ and $\bv$. Let $A_\Delta$ and $\bv_\Delta$ denote changes in $A$ and $\bv$. Define $A_n = A_o + A_\Delta$ and $\bv_n = \bv_o + \bv_\Delta$. We can compute the updated attention-value product $A_n \bv_n$ as
\begin{align}
    A_n \bv_n
        &= (A_o + A_\Delta) (\bv_o + \bv_\Delta) \nonumber \\
        &= A_o \bv_o + A_o \bv_\Delta + A_\Delta \bv_o + A_\Delta \bv_\Delta \nonumber \\
        &= A_o \bv_o + (A_o + A_\Delta) \bv_\Delta + A_\Delta (\bv_o + \bv_\Delta) - A_\Delta \bv_\Delta \nonumber \\
        &= A_o \bv_o + A_n \bv_\Delta + A_\Delta \bv_n - A_\Delta \bv_\Delta. \label{eq:delta}
\end{align}
Therefore, on each time step, we can update $A \bv$ by adding $A_n \bv_\Delta + A_\Delta \bv_n - A_\Delta \bv_\Delta$ to the previous result $A_o \bv_o$.

We obtain $A_\Delta$, $\bv_\Delta$, $A_o$, and $\bv_o$ using \emph{delta gate modules}. Delta gates are similar to the gates defined in Section~\ref{sec:gating}, with one difference: instead of returning $\tbc$, a delta gate returns $\bu$ and $\tbe$ (where $\tbe$ is the result of gathering the selected indices from $\be$). $\bu$ represents the effective current value of the gate's output, corresponding to $A_n$ or $\bv_n$ in Equation~\ref{eq:delta}. $\tbe$ represents the amount of change on the current time step, corresponding to $A_\Delta$ or $\bv_\Delta$ in Equation~\ref{eq:delta}.

\begin{figure}
    \centering
    \includegraphics{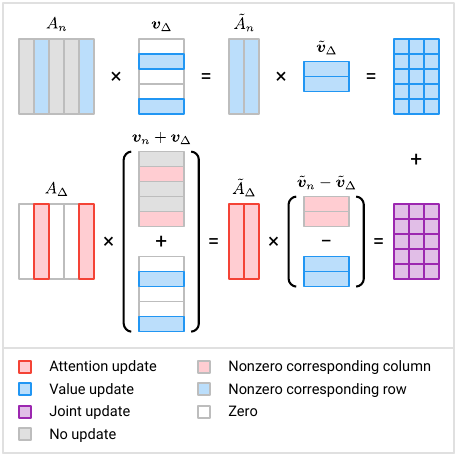}
    \caption{\textbf{The attention-value product.} We propose a delta-based strategy for sparsely updating the product $A \bv$. We reduce the cost of each sub-product by cutting rows and columns that do not contribute to the result (due to a zero multiplication).}
    \label{fig:matmul2}
\end{figure}

Figure~\ref{fig:matmul2} illustrates our approach for efficiently computing the three delta terms in Equation~\ref{eq:delta}. We remove the columns of $A_n$ that correspond to zero rows in $\bv_\Delta$ (these columns will always be multiplied by zero). Let $\tA_n$ denote $A_n$ with these columns removed. We remove rows of $\bv_n$ analogously to produce $\tbv_n$. We then compute
\begin{equation}
    \tA_n \tbv_\Delta + \tA_\Delta \tbv_n - \tA_\Delta \tbv_\Delta, \label{eq:matmul2}
\end{equation}
adding the result to the previous value of $A \bv$.

The product $\tA_\Delta \tbv_\Delta$ assumes the columns of $\tA_\Delta$ are correctly aligned with the rows of $\tbv_\Delta$. We achieve this alignment by forcing the $A$ gate to select the same indices as the $\bv$ gate. Using a separate policy in the $A$ gate would be possible, but would require a re-alignment operation before computing $\tA_\Delta \tbv_\Delta$. Further, forcing alignment allows us to eliminate a multiplication by rearranging Equation~\ref{eq:matmul2} as
\begin{equation}
    \tA_n \tbv_\Delta + \tA_\Delta (\tbv_n - \tbv_\Delta). \label{eq:matmul2_aligned}
\end{equation}
Equation~\ref{eq:matmul2_aligned} has a cost of $2 M N D$ (assuming the addition has a negligible cost), compared to $N^2 D$ for a standard multiplication. We see savings when $M < N / 2$.

\subsection{Token Selection Policies}
\label{sec:policies}
An important design choice for an Eventful Transformer is the token selection policy. Given a gate error tensor $\be$, a policy generates a mask $\bm{m}$ indicating which tokens should be updated. We now discuss the design of selection policies.

\paragraphskip
\paragraphtitle{Top-$\bm{r}$ policy.} This policy selects the $r$ tokens whose error $\be$ has the largest norm (we use the L2 norm). The top-$r$ policy is lightweight and has a single parameter that can be easily tuned by hand. Varying $r$ gives direct control over the model's computation cost. These properties make the top-$r$ policy a good fit for applications with tight (potentially time-varying) computational constraints. We use a top-$r$ policy in our main experiments.

\paragraphskip
\paragraphtitle{Threshold policy.} This policy selects all tokens where the norm of the error $\be$ exceeds a threshold $h$. A threshold policy is input-adaptive; the number of tokens selected depends on the amount of change in the scene. This input adaptivity can potentially lead to a better accuracy-cost tradeoff. However, the best value for the threshold $h$ depends on the distribution of token vectors (which varies across layers) and is difficult to decide. In addition, a threshold policy does not give a fixed compute cost. This policy is likely better suited to applications with more flexible resources, where achieving the best possible accuracy-cost tradeoff is critical.

\paragraphskip
\paragraphtitle{Other policies.} More sophisticated token selection policies could lead to an improved accuracy-cost tradeoff. For example, we could use a learned policy (\eg, a lightweight policy network). However, training the policy's decision mechanism might be challenging, due to the general non-differentiability of the binary mask $\bm{m}$. Another idea is to use an importance score (\eg, as proposed in \cite{fayyazAdaptiveToken2022}) to inform the selection. We leave these ideas as potential topics for future work.

\section{Experiments}
\label{sec:experiments}

In this section, we present our experiments and results. We evaluate our method for video object detection (Section~\ref{sec:exp_detection}) and video action recognition (Section~\ref{sec:exp_action}). We show additional analysis in Section~\ref{sec:exp_others}.

\subsection{Video Object Detection}
\label{sec:exp_detection}

\begin{figure}
    \centering
    \includegraphics{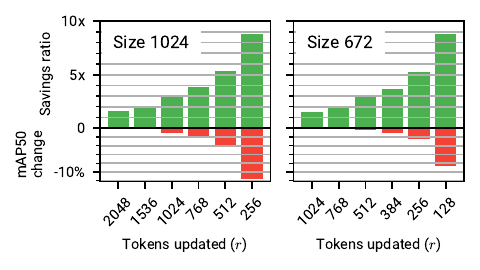}
    \caption{\textbf{Video object detection results.} Computation savings ratio (positive axis) and relative reductions in mAP50 score (negative axis) for our method. Results are for the ViTDet model~\cite{liExploringPlain2022} on the VID~\cite{russakovsky2015ImageNet} dataset. See the supplement for tables.}
    \label{fig:computation_savings}
\end{figure}

\paragraphtitle{Task and dataset.} We test our method on video object detection using the ILSVRC 2015 ImageNet VID dataset~\cite{russakovsky2015ImageNet}. We report results on the validation set, which contains 555 videos with lengths of up to 2895 frames. Following prior works~\cite{chdnMemoryEnhanced2020,redmonYouOnly2016}, we evaluate the mean average precision (mAP) metric with an IoU threshold of 0.5.

\paragraphskip
\paragraphtitle{Implementation details.} We consider the ViTDet model from~\cite{liExploringPlain2022}, which we apply to individual frames of an input video. ViTDet combines a plain Transformer backbone (based on ViT-B \cite{dosovitskiyImageWorth2020}) with a standard detection head~\cite{caiCascadeRCNN2021,heMaskRCNN2017}. The backbone consists of 12 blocks with interleaved global and windowed self-attention (blocks 3, 6, 9, and 12 use global attention). Windowed self-attention uses a window size of 14$\times$14 tokens (224$\times$224 pixels). Token vectors are 768-dimensional. Self-attention operators have 12 heads and employ learned relative position embeddings.

Before the backbone, the model maps each 16$\times$16 image patch to a token vector using a linear transform. The model expects fixed-size inputs (due to resolution-specific position embeddings). Therefore, following from~\cite{liExploringPlain2022}, we rescale and pad all video frames to a uniform size (\eg, 1024$\times$1024) before applying the model.

We convert the model to an Eventful Transformer following the method in Section~\ref{sec:methods}. In blocks that use windowed attention, we exploit temporal redundancy only within token-wise operations (not within the query-key or attention-value products). Our complete approach is compatible with windowed attention; however, windowing leads to some implementation challenges (ragged tensor shapes across windows, making batched computation more difficult). Note that for ViTDet, global self-attention represents the bulk of the self-attention compute cost.

\begin{figure}
    \centering
    \includegraphics{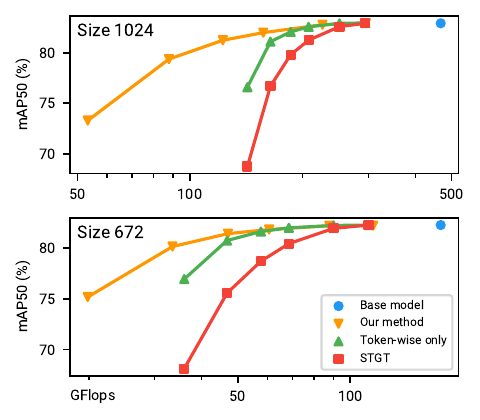}
    \caption{\textbf{Video object detection comparison and ablation.} The accuracy-cost tradeoff for our method, compared with STGT~\cite{liSpatiotemporalGated2021} and an ablation that only accelerates token-wise operations. See the supplement for tables.}
    \label{fig:vitdet}
\end{figure}

\paragraphskip
\paragraphtitle{Experiment protocol and baselines.} We fine-tune the original ViTDet weights (trained on COCO) for VID object detection. See the supplement for training parameters. Note that we fine-tune \emph{before} we add temporal redundancy awareness to the model. We train and evaluate at resolution 1024$\times$1024. To understand the effect of token count (which strongly influences compute cost), we also evaluate at resolution 672$\times$672. Rather than training a separate lower-resolution model, we adapt the 1024$\times$1024 model by interpolating the learned position embeddings. The resulting adapted model retains most of its accuracy.

We compare against a version of the STGT method~\cite{liSpatiotemporalGated2021}. Due to unavailable source code, we were unable to evaluate all components of this method (notably, the use of a learned policy network). Instead, we consider a simplified version that uses the same top-$r$ policy as our method. This setup enables a direct comparison of the core gating and update mechanisms. In addition, we evaluate an ablated version of our approach that only accelerates token-wise operations. We vary the policy $r$ to explore the accuracy-compute tradeoff. At resolution 1024, we test $r=$~256, 512, 768, 1024, 1536, and 2048 (from a maximum of 4096 tokens). At resolution 672, we test $r=$~128, 256, 384, 512, 768, and 1024 (from a maximum of 1764).

\paragraphskip
\paragraphtitle{Results.} Figure~\ref{fig:computation_savings} shows our results. Our method gives significant savings with only minor reductions in accuracy. For example, at size 1024 with $r=$~768, our approach reduces the cost from 467.4 GFlops to 122.3 GFlops (3.8x lower) while reducing the mAP50 score from 82.93 to 81.25 (-1.68\% in absolute mAP50). At size 672 with $r=$~384, we reduce the cost by 3.7x with a -0.85\% change in mAP50.

In these experiments, some tokens correspond to padded space and are therefore ``easy'' from a temporal redundancy standpoint. However, even in a padding-free deployment (\eg, with a single, known training and inference resolution) our method would still give strong computation savings. For example, consider resolution 1024 with $r = 768$. We are skipping 66\% of all non-padded tokens here (based on a measured mean padding ratio of 44.6\% on VID -- corresponding to a $\sim$16:9 aspect ratio). This corresponds to a savings of $>$2x, with an accuracy drop of only 1.68\%. Note that our ViViT experiments (Section~\ref{sec:exp_action} and supplementary) do not involve padding.

Figure~\ref{fig:vitdet} shows the accuracy-compute tradeoff for our method, along with baselines. Our approach gives a considerable improvement in the accuracy-compute tradeoff compared to STGT~\cite{liSpatiotemporalGated2021}. Further, adding redundancy awareness to the query-key and attention-value products reduces the cost significantly, especially at low $r$ values.

\subsection{Video Action Recognition}
\label{sec:exp_action}

\begin{figure}
    \centering
    \includegraphics{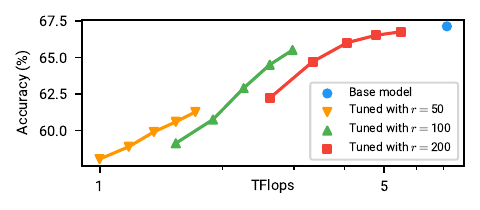}
    \caption{\textbf{Video action recognition results.} Our results for action recognition on EPIC-Kitchens 100 using the ViViT model \cite{arnabViViTVideo2021}. We report the total TFlops per video (spatial + temporal sub-models). See the supplement for a table containing this data.}
    \label{fig:vivit}
\end{figure}

\begin{table}
    \centering
    \caption{\textbf{Adding spatial redundancy to ViTDet.} ``Spatial'' is a model with pooling in $\bk$ and $\bv$. ``Spatiotemporal'' is a model with both pooling and temporal redundancy awareness.}
    \tableskip
    \label{tab:spatial_1024}
    \small
    \begin{tabular}{lccc}
        \toprule
        Variant        & $r$  & mAP50 (\%) & GFlops \\ \midrule
        Base model     & --   & 82.93      & 467.4  \\
        Spatial        & --   & 80.15      & 388.1  \\
        Spatiotemporal & 2048 & 80.14      & 217.0  \\
        Spatiotemporal & 1536 & 80.07      & 169.3  \\
        Spatiotemporal & 1024 & 79.50      & 121.0  \\
        Spatiotemporal & 768  & 78.69      & 96.3   \\
        Spatiotemporal & 512  & 76.96      & 70.9   \\
        Spatiotemporal & 256  & 71.35      & 44.5   \\
        \bottomrule
    \end{tabular}
\end{table}

\paragraphtitle{Task and dataset.} We evaluate our method on action recognition using the EPIC-Kitchens 100 dataset~\cite{damenScalingEgocentric2018}. EPIC-Kitchens 100 contains highly dynamic egocentric videos annotated with 97 verb and 300 noun classes. We consider the verb classification task. The training and validation set contains 67217 and 9668 action instances, respectively.

\paragraphskip
\paragraphtitle{Implementation details.} We use the ViViT model~\cite{arnabViViTVideo2021} with factorized spatial and temporal sub-models based on ViT-B. The spatial sub-model (the bulk of the compute cost) is applied sequentially to 16 2-frame input clips. The outputs of the spatial model are concatenated and passed to the temporal model, which returns a class prediction. The prediction is the average over 12 video views (4 temporal views, each divided into 3 spatial crops). Each view has a shape of 320$\times$320$\times$32. Unlike ViTDet, ViViT adds a class embedding token (see~\cite{dosovitskiyImageWorth2020}), does not use windowed self-attention, and does not use relative position embeddings.

We convert the spatial model to an Eventful Transformer. Naively replacing the spatial model with an Eventful version leads to a considerable drop in accuracy (about -10\% with $r=$~100). We conjecture that the cause is a distribution shift in the inputs to the temporal model (see the supplement for further discussion). We recover most of the lost accuracy by fine-tuning the \emph{non-Eventful} temporal model on the outputs of a frozen Eventful spatial model.

\paragraphskip
\paragraphtitle{Experiment protocol.} We start with ViViT pre-trained on EPIC-Kitchens 100 and fine-tune the temporal model as described above (on the EPIC-Kitchens training set). We fine-tune different model variants with policy $r$ values of 50, 100, and 200 (out of a maximum of 401 tokens). See the supplement for training parameters. We report results using the top-1 accuracy metric, following standard protocol~\cite{damenScalingEgocentric2018}.

\paragraphskip
\paragraphtitle{Results.} Figure~\ref{fig:vivit} shows our results for the Eventful ViViT model. We evaluate a range of $r$ values for each of the fine-tuned variants. We test the original fine-tuned $r$-value, along with $\pm$20\% and $\pm$40\% of this value. We observe considerable computation savings with only moderate reductions in accuracy. For example, with $r =$~140, we reduce the cost by 2.4x while reducing the accuracy by only 1.62\%. In addition, the model retains adaptivity despite being fine-tuned with a single-$r$ value, exhibiting a favorable accuracy-compute tradeoff over a range of $r$-values.

\subsection{Spatial Redundancy and Runtime}
\label{sec:exp_others}

\paragraphtitle{Considering spatial redundancy.} Eventful Transformers exploit \emph{temporal} redundancy and thus complement prior works that leverage \emph{spatial} redundancy. Here we present a simple proof-of-concept experiment that considers spatial redundancy in Eventful Transformers.

Specifically, we adopt a variant of~\cite{xiongNystromformerNystromBased2021}, which applies spatial pooling to the self-attention key and value tokens. We apply this method with 2$\times$2 pooling to the global self-attention operators in the ViTDet model. We evaluate this method both with and without temporal redundancy awareness. In the temporal-redundancy model, we pool $\bk$ and $\bv$ after their respective buffers. We pool $\tbk$ by first pooling the active indices (equivalent to max-pooling the mask $\bm{m}$), then gathering the buffered $\bk$ using the pooled indices. 

Table~\ref{tab:spatial_1024} shows our results for resolution 1024 (see the supplement for resolution 672). We see that the spatial and temporal methods are complementary; both meaningfully contribute to reducing the computational cost. See Section~\ref{sec:discussion} for further discussion on spatial redundancy methods.

\paragraphskip
\paragraphtitle{Runtime.} We show preliminary runtime results on a CPU (Xeon Silver 4214, 2.2 GHz) and a GPU (NVIDIA RTX 3090). See the supplementary material for experiment details. Table~\ref{tab:runtime} shows our results. Adding temporal redundancy awareness leads to speedups of up to 1.74x on the GPU and 2.48x on the CPU. These results should be seen just as a proof of concept -- we are confident that these speedups could be improved with further engineering effort (\eg, by replacing vanilla PyTorch operators with custom kernels or using a high-performance inference framework).

\begin{table}
    \centering
    \caption{\textbf{Runtimes (ms).} ViTDet runtimes are for the Transformer backbone only. ViViT runtimes include the temporal sub-model.}
    \tableskip
    \label{tab:runtime}
    \small
    \begin{tabular}{lclccc}
        \toprule
        Model  & Size & Variant        & $r$ & GPU  & CPU    \\ \midrule
        ViTDet & 1024 & Base model     & --  & 86.6 & 5150   \\
        ViTDet & 1024 & Spatial        & --  & 58.9 & 3116   \\
        ViTDet & 1024 & Temporal       & 512 & 69.9 & 3570   \\
        ViTDet & 1024 & Spatiotemporal & 512 & 38.1 & 1682   \\ \midrule
        ViTDet & 672  & Base model     & --  & 28.3 & 1492   \\
        ViTDet & 672  & Spatial        & --  & 23.3 & 1055   \\
        ViTDet & 672  & Temporal       & 256 & 21.6 & 838    \\
        ViTDet & 672  & Spatiotemporal & 256 & 20.8 & 478    \\ \midrule
        ViViT  & 320  & Base model     & --  & 950  & 5.45e4 \\
        ViViT  & 320  & Temporal       & 50  & 545  & 2.15e4 \\
        \bottomrule
    \end{tabular}
\end{table}

\begin{figure}
    \centering
    \includegraphics{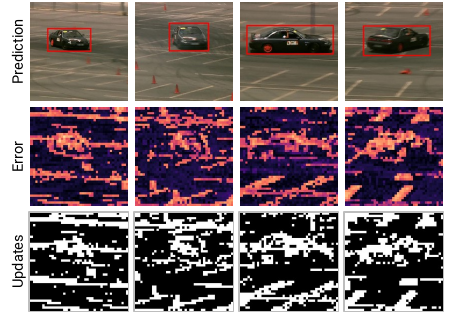}
    \caption{\textbf{Visualization of updates.} The error $\be$ and update mask $\bm{m}$, for the pre-QKV gate in block 3 of ViTDet. Video source: \cite{russakovsky2015ImageNet}}
    \label{fig:updates}
\end{figure}

\paragraphskip
\paragraphtitle{Visualization of updates.} Figure~\ref{fig:updates} shows an example video sequence. We visualize the model predictions (top), the token-wise L2 norm of the error $\be$ (middle), and the update mask $\bm{m}$ (bottom). We see that larger error values correspond to dynamic regions in the image.

\section{Discussion}
\label{sec:discussion}

\paragraphtitle{Memory overhead.} Our method reduces floating point operations at the cost of higher memory usage. Each gate or buffer maintains a reference tensor ($\bu$ or $\bb$, respectively). The memory overhead for token gates and buffers is generally modest. For, consider size-1024 ViTDet. The model has 4096 768-dimensional tokens, meaning a token gate or buffer takes 12.6/6.3~MB of memory at full/half precision.

However, gating or buffering the attention matrix $A$ can require a larger amount of memory. For example, in the global attention layers of the size-1024 ViTDet model, the $A$ matrix has shape 4096$\times$4096$\times$12. Buffering this $A$ requires 805/403~MB at full/half precision. Fortunately, the situation dramatically improves if we reduce the number of tokens or use windowed attention (due to the quadratic size of $A$). For example, the $A$ matrices for the size-672 ViTDet model (1764 tokens) and the ViViT model (301 tokens) occupy 149/75~MB and 4.3/2.2~MB, respectively. In applications where memory is tight and the $A$ matrix is large, it is possible to save memory by removing temporal redundancy awareness in the query-key and/or attention-value products (each eliminates one $A$-shaped state tensor).

\paragraphskip
\paragraphtitle{Integration with spatial redundancy methods.} A promising avenue for future work the is further integration of our approach with spatial redundancy methods. Conceptually, these methods summarize a large, redundant set of tokens using a more compact set of tokens. The gating module in an Eventful Transformer assumes that most of its inputs vary smoothly over time. When combining our approach with spatial redundancy methods, we need to ensure that the compact spatial summary is relatively stable. For example, with adaptive token clustering \cite{bolyaTokenMerging2022,marinTokenPooling2021}, we would need to sort the clusters in a mostly-consistent order.

There are many potentially interesting questions regarding joint modeling of spatial and temporal redundancy. For example, how is the temporal axis different from the spatial one? Should the temporal axis be modeled separately? We leave such questions for future work.

\paragraphskip
\paragraphtitle{Acknowledgments.} This research was supported in part by NSF CAREER award \#1943149.

{
    \small
    \bibliographystyle{ieee_fullname}
    \bibliography{full}
}

\clearpage

\renewcommand{\theequation}{\Alph{equation}}
\renewcommand{\thefigure}{\Alph{figure}}
\renewcommand{\thesection}{\Alph{section}}
\renewcommand{\thetable}{\Alph{table}}

\setcounter{equation}{0}
\setcounter{figure}{0}
\setcounter{section}{0}
\setcounter{table}{0}

\twocolumn[
    \centering
    \vspace{0.5in}
    {\Large \textbf{Supplementary Material}}
    \vspace{0.75in}
]

We use capital letters (\eg, Figure~A) to refer to the supplementary material and numbers (\eg, Figure~1) to refer to the main paper.

In Section~\ref{sec:further_discussion}, we provide further discussion on token selection policies, optimizations to the query-key product, and the ViViT temporal model. In Section~\ref{sec:additional_experiments}, we present additional experiments: action recognition on Kinetics-400, an evaluation of a threshold policy, and an ablation of the gate position. In Section~\ref{sec:experiment_details}, we provide low-level details for the experiments in the main paper. In Section~\ref{sec:result_tables}, we include tables of results for the experiments in the main paper.

\section{Further Discussion}
\label{sec:further_discussion}

\paragraphtitle{The ViViT temporal sub-model.} Recall that, for ViViT action recognition, we fine-tune the non-Eventful temporal model on the outputs of the Eventful spatial model. We now provide some intuition as to why this is necessary to preserve the prediction accuracy.

The outputs of an Eventful Transformer are approximations of the ``correct'' outputs (those of the original, non-Eventful Transformer). In the case of the ViViT spatial model, individual outputs are fairly close to the correct values. However, the \emph{pattern of temporal changes} between outputs may be quite different from the original model. Token gates reduce the number of updated tokens on each frame, but each update tends to be larger (a single update may contain accumulated changes from several time steps). Given the nature of the prediction task -- action recognition on highly dynamic videos -- the temporal sub-model is sensitive to the pattern of temporal changes. Fine-tuning allows us to correct for the shifts in these temporal changes that result from using an Eventful spatial model.

\paragraphskip
\paragraphtitle{Compatibility with spatial redundancy methods.} We now provide further discussion regarding the compatibility of our method with spatial redundancy approaches. Abstractly, we can think of spatial redundancy methods as summarizing a set of tokens $\bx \in \R^{N \times D}$ using a reduced set of tokens $\hat{\bx} \in \R^{M \times D}$. The simple method in our experiments summarizes tokens using uniform pooling; however, we could also use adaptive pruning or merging.

Assume we apply a gate to the reduced tokens $\hat{\bx}$. The gate assumes that the definitions of its input tokens are relatively stable. This assumption clearly holds for non-reduced or uniformly pooled tokens. However, we need to be careful when applying arbitrary reductions to $\bx$.

For example, say we have an image containing a region of blue sky. An adaptive token merging method might combine all sky-colored tokens from $\bx$ into a single token in $\hat{\bx}$. Assume that on frame $t = 1$, the first token in $\hat{\bx}$ represents the sky. Ideally, on frame $t = 2$, the first token in $\hat{\bx}$ should again represent the sky. Note that this is not a strict constraint -- our gating logic can deal with non-consistent definitions for a few tokens. However, if the definitions for all tokens in $\hat{\bx}$ completely change between frames, then the gate will not be able to keep up (\ie, the number of tokens with significant changes will exceed the policy $r$-value).

\section{Additional Experiments}
\label{sec:additional_experiments}

\paragraphtitle{Video action recognition on Kinetics-400.} We evaluate our method on the Kinetics-400 action recognition dataset~\cite{kayKineticsHuman2017}. Kinetics-400 contains over 300k video clips, each annotated with one of 400 action categories. We evaluate top-1 accuracy. We use the same ViViT model architecture as in our EPIC-Kitchens experiments; the only difference is the input size (224$\times$224 rather than 320$\times$320).

As in our EPIC-Kitchens experiments, we fine-tune the non-Eventful temporal model on the outputs of the Eventful spatial model. We fine-tune three variants of the model with $r=$24, 48, and 96 (out of a maximum of 197 tokens). We train for 10 epochs on a subset of the training set containing 39729 videos. We use the AdamW optimizer~\cite{loschilovDecoupledWeight2019} with a learning rate of 2$\times$10\textsuperscript{-6}, weight decay of 0.05, and a batch size of 16 videos. We add 50\% dropout before the final classification layer.

Table~\ref{tab:kinetics} shows our results. The accuracy-compute tradeoff is generally consistent with our results on EPIC-Kitchens. For example, with $r=$~96, we sacrifice 1.48\% accuracy for a speedup of approximately 2x.

\begin{table}
    \centering
    \caption{\textbf{Kinetics-400 video action recognition.} Results for Kinetics-400 action recognition using the ViViT model. We report the total TFlops per video (spatial + temporal sub-models).}
    \tableskip
    \label{tab:kinetics}
    \small
    \begin{tabular}{lccc}
        \toprule
        Variant    & $r$ & Accuracy (\%) & TFlops \\ \midrule
        Base model & --  & 79.06         & 3.360  \\
        Temporal   & 96  & 77.62         & 1.814  \\
        Temporal   & 48  & 75.88         & 1.016  \\
        Temporal   & 24  & 75.16         & 0.618  \\
        \bottomrule
    \end{tabular}
\end{table}

\paragraphskip
\paragraphtitle{A threshold policy.} We evaluate the ViTDet object detection model with a threshold policy. The threshold policy selects all tokens where the L2 norm of $\be$ exceeds a threshold $h$. We test $h=$~0.2, 1.0, and 5.0. See Table~\ref{tab:threshold_policy} for results. The accuracy-compute tradeoff for the threshold policy is generally worse than for the top-$r$ policy. For example, compare threshold $h=$~5.0 with $r=$~512 in Table~\ref{tab:vitdet}. This is likely due to the use of a constant threshold for all gates (we would ideally use a unique threshold for each gate).

\section{Experiment Details}
\label{sec:experiment_details}

\paragraphtitle{Fine-tuning ViTDet for VID.} We initialize our model using COCO~\cite{linMicrosoftCOCO2014} pre-trained weights, and then trained on a combination of the ImageNet VID and ImageNet DET datasets, following common protocols in~\cite{chdnMemoryEnhanced2020,zhuFlowGuided2017}. We select images from the DET dataset that are of of the same 30 classes as in the VID dataset. The training uses a batch size of 8, a maximum input resolution of 1024$\times$1024, an initial learning rate of 10\textsuperscript{-4}, and a weight decay of 0.1. We use the AdamW optimizer~\cite{loschilovDecoupledWeight2019} with linear warmup for a total of 5 epochs, with 10x learning rate decay from the 3rd epoch.

\paragraphskip
\paragraphtitle{Fine-tuning the ViViT temporal model.} We fine-tune the temporal sub-model for 5 epochs. We use the AdamW optimizer~\cite{loschilovDecoupledWeight2019} with a learning rate of 10\textsuperscript{-5}, weight decay of 0.05, and a batch size of 8 videos. We add 50\% dropout before the final classification layer.

\paragraphskip
\paragraphtitle{Arithmetic precision.} We compute the product $A \bv$ at half precision in the global self-attention operators of the Eventful model. Using half precision reduces the model's computational cost and memory footprint and has a negligible effect on accuracy. When evaluating runtimes, we also compute $A \bv$ at half precision in the base model (this ensures a fair comparison).

\paragraphskip
\paragraphtitle{Runtime experiments.} For ViTDet, we evaluate CPU runtimes using one random video from VID (ID 00023010, containing 242 frames). On the GPU, we use 5 random videos. For ViViT, we evaluate CPU runtimes using 5 random videos from EPIC-Kitchens. On the GPU, we use 100 random videos. We use a consistent random seed across all experiment runs.

\paragraphskip
\paragraphtitle{Operation counting.} Our GFlop counts include the following types of operations: linear transforms, matrix multiplications, einsum operations (used in relative position embeddings), and additions. We count a multiply-accumulate as a single operation. In Eventful Transformers, we additionally count operations required for updating the gate (additions and subtractions) and the extra additions in the sparse attention-value update. We only report operations in the Transformer backbones (\eg, we do not count anything in the object detection head).

\section{Result Tables}
\label{sec:result_tables}

In this section, we provide tables of results for experiments in the main paper. Table~\ref{tab:vitdet} corresponds to Figures~7 and 8, and Table~\ref{tab:vivit} corresponds to Figure~9. Table~\ref{tab:spatial_672} shows spatial redundancy results for the 672-resolution ViTDet model (the 1024-resolution results are in Table~1).

\begin{table}
    \centering
    \caption{\textbf{A threshold policy.} Results for a threshold policy with the 1024-resolution ViTDet model. The policy selects tokens where the error $\be$ exceeds a threshold $h$.}
    \tableskip
    \label{tab:threshold_policy}
    \small
    \begin{tabular}{lccc}
        \toprule
        Variant    & $h$ & mAP50 (\%) & GFlops \\ \midrule
        Base model & --  & 82.93      & 467.4  \\
        Temporal   & 0.2 & 83.00      & 431.8  \\
        Temporal   & 1.0 & 82.75      & 294.1  \\
        Temporal   & 5.0 & 78.11      & 133.5  \\
        \bottomrule
    \end{tabular}
\end{table}

\begin{table}
    \centering
    \caption{\textbf{Video object detection results.} Results for video object detection on VID using the ViTDet model. This table corresponds to Figures~7 and 8 in the main paper.}
    \tableskip
    \label{tab:vitdet}
    \small
    \begin{tabular}{clccc}
        \toprule
        Size & Variant         & $r$  & mAP50 (\%) & GFlops \\ \midrule
        1024 & Base model      & --   & 82.93      & 467.4  \\
        1024 & Our method      & 2048 & 82.94      & 294.9  \\
        1024 & Our method      & 1536 & 82.79      & 225.9  \\
        1024 & Our method      & 1024 & 82.00      & 156.8  \\
        1024 & Our method      & 768  & 81.25      & 122.3  \\
        1024 & Our method      & 512  & 79.38      & 87.8   \\
        1024 & Our method      & 256  & 73.29      & 53.3   \\
        1024 & Token-wise only & 2048 & 82.97      & 294.1  \\
        1024 & Token-wise only & 1536 & 82.93      & 250.7  \\
        1024 & Token-wise only & 1024 & 82.58      & 207.3  \\
        1024 & Token-wise only & 768  & 82.08      & 185.7  \\
        1024 & Token-wise only & 512  & 81.11      & 164.0  \\
        1024 & Token-wise only & 256  & 76.60      & 142.3  \\
        1024 & STGT            & 2048 & 82.92      & 294.1  \\
        1024 & STGT            & 1536 & 82.60      & 250.7  \\
        1024 & STGT            & 1024 & 81.25      & 207.3  \\
        1024 & STGT            & 768  & 79.81      & 185.7  \\
        1024 & STGT            & 512  & 76.70      & 164.0  \\
        1024 & STGT            & 256  & 68.73      & 142.3  \\ \midrule
        672  & Base model      & --   & 82.28      & 174.5  \\
        672  & Our method      & 1024 & 82.23      & 115.1  \\
        672  & Our method      & 768  & 82.21      & 87.9   \\
        672  & Our method      & 512  & 81.84      & 60.7   \\
        672  & Our method      & 384  & 81.43      & 47.1   \\
        672  & Our method      & 256  & 80.16      & 33.5   \\
        672  & Our method      & 128  & 75.19      & 19.9   \\
        672  & Token-wise only & 1024 & 82.28      & 111.9  \\
        672  & Token-wise only & 768  & 82.25      & 90.2   \\
        672  & Token-wise only & 512  & 82.01      & 68.5   \\
        672  & Token-wise only & 384  & 81.64      & 57.7   \\
        672  & Token-wise only & 256  & 80.76      & 46.8   \\
        672  & Token-wise only & 128  & 76.96      & 36.0   \\
        672  & STGT            & 1024 & 82.28      & 111.9  \\
        672  & STGT            & 768  & 81.95      & 90.2   \\
        672  & STGT            & 512  & 80.45      & 68.5   \\
        672  & STGT            & 384  & 78.71      & 57.7   \\
        672  & STGT            & 256  & 75.57      & 46.8   \\
        672  & STGT            & 128  & 68.13      & 36.0   \\
        \bottomrule
    \end{tabular}
\end{table}

\begin{table}
    \centering
    \caption{\textbf{Video action recognition results.} Results for video action recognition on EPIC-Kitchens using the ViViT model. This table corresponds to Figure~9 in the main paper.}
    \tableskip
    \label{tab:vivit}
    \small
    \begin{tabular}{lcccc}
        \toprule
        Variant    & Tuned $r$ & Tested $r$ & Accuracy (\%) & TFlops \\ \midrule
        Base model & --        & --         & 67.14         & 7.12   \\
        Temporal   & 200       & 280        & 66.77         & 5.49   \\
        Temporal   & 200       & 240        & 66.53         & 4.77   \\
        Temporal   & 200       & 200        & 66.02         & 4.05   \\
        Temporal   & 200       & 160        & 64.72         & 3.33   \\
        Temporal   & 200       & 120        & 62.23         & 2.62   \\
        Temporal   & 100       & 140        & 65.52         & 2.98   \\
        Temporal   & 100       & 120        & 64.51         & 2.62   \\
        Temporal   & 100       & 100        & 62.91         & 2.26   \\
        Temporal   & 100       & 80         & 60.76         & 1.90   \\
        Temporal   & 100       & 60         & 59.13         & 1.54   \\
        Temporal   & 50        & 70         & 61.27         & 1.72   \\
        Temporal   & 50        & 60         & 60.60         & 1.54   \\
        Temporal   & 50        & 50         & 59.91         & 1.36   \\
        Temporal   & 50        & 40         & 58.90         & 1.18   \\
        Temporal   & 50        & 30         & 58.05         & 1.00   \\
        \bottomrule
    \end{tabular}
\end{table}

\begin{table}
    \centering
    \caption{\textbf{Adding spatial redundancy to 672-resolution ViTDet.} Results for adding spatial redundancy to the 672-resolution ViTDet model. 1024-resolution results are in the main paper.}
    \tableskip
    \label{tab:spatial_672}
    \small
    \begin{tabular}{lccc}
        \toprule
        Variant        & $r$  & mAP50 (\%) & GFlops \\ \midrule
        Base model     & --   & 82.28      & 174.5  \\
        Spatial        & --   & 79.86      & 159.7  \\
        Spatiotemporal & 1024 & 79.85      & 98.2   \\
        Spatiotemporal & 768  & 79.81      & 75.5   \\
        Spatiotemporal & 512  & 79.47      & 52.8   \\
        Spatiotemporal & 384  & 79.02      & 41.4   \\
        Spatiotemporal & 256  & 77.90      & 29.8   \\
        Spatiotemporal & 128  & 73.40      & 18.0   \\
        \bottomrule
    \end{tabular}
\end{table}

\end{document}